\documentclass[conference]{IEEEtran}
\IEEEoverridecommandlockouts
\usepackage{cite}
\usepackage{amsmath,amssymb,amsfonts}
\usepackage{algorithmic}
\usepackage{graphicx}
\usepackage{textcomp}
\usepackage{xcolor}
\usepackage{dsfont}
\def\BibTeX{{\rm B\kern-.05em{\sc i\kern-.025em b}\kern-.08em
    T\kern-.1667em\lower.7ex\hbox{E}\kern-.125emX}}

\newcommand{\add}[1]{#1}

\begin{document}

\title{Fuzzy Conceptual Graphs: a comparative discussion}

\author{\IEEEauthorblockN{Adam Faci}
\IEEEauthorblockA{\textit{LIP6, Sorbonne Université, CNRS}\\
F-75005 Paris, France\\
\textit{THALES}, 91477 Palaiseau, France \\
adam.faci@lip6.fr}
\and
\IEEEauthorblockN{Marie-Jeanne Lesot}
\IEEEauthorblockA{\textit{LIP6, Sorbonne Université, CNRS}\\
F-75005 Paris, France\\
marie-jeanne.lesot@lip6.fr}
\and
\IEEEauthorblockN{Claire Laudy}
\IEEEauthorblockA{\textit{THALES}, 91477 Palaiseau, France \\
claire.laudy@thalesgroup.com}
}

\maketitle

\begin{abstract}
Conceptual Graphs (CG) are a graph-based knowledge representation and reasoning formalism; fuzzy Conceptual Graphs (fCG) constitute an extension that enriches their expressiveness, exploiting the fuzzy set theory so as to relax their constraints at various levels. This paper proposes a comparative study of existing approaches over their respective advantages and possible limitations. The discussion revolves around three axes: (a) Critical view of {each approach} and comparison with previous propositions from the state of the art; (b) Presentation of the many possible interpretations of \add{each} definition to illustrate its potential and its limits; (c) {Clarification} of the part of CG impacted by the definition as well as the relaxed constraint.
\end{abstract}


\section{Introduction}

Conceptual Graphs (CG)~\cite{chein_conceptual_2008} are a graph-based knowledge representation family of formalisms whose specificities are a clear distinction between ontological knowledge and factual knowledge, reasoning mechanisms based on graph operations and foundations in first-order logic.

Their expressiveness has been extended by the addition or relaxation of constraints. For instance, inference rules and constraints have been introduced \add{resulting in} different formalisms of CG~\cite{baget2002extensions}; and datatypes have been proposed to represent non-symbolic values~\cite{baget2007datatype}.

This paper focuses on propositions that enable the representation and manipulation of imprecise knowledge, \add{in particular in the framework of the fuzzy set theory~\cite{zadeh1965,zimmermann2011fuzzy}.} These propositions broaden the CG's knowledge representation capacity through the definition of weights or the embedding of fuzzy sets, that quantify the imprecision. These extensions occur in either the ontological or the factual part. They have been exploited in different application domains, such as image matching~\cite{mulhemy2001fuzzy}, database querying~\cite{thomopoulos2006fuzzy} or ontology mapping~\cite{buche2008ontology}.

The goal of this paper is to \add{review the} state of the art of the different methods used to represent imprecise information in CG in order to extend their knowledge representation capacity. In particular, it examines the constraints on the CG formalism, by highlighting which constraint is relaxed, how it is relaxed and what are the consequences in terms of knowledge representation. 

The paper focuses on knowledge representation and does not consider the question of reasoning mechanisms. \add{Indeed, it is essential to examine first the knowledge representation formalisms and distinguish fundamental differences before even considering associated reasoning elements.} Section~\ref{sec:edla} provides a review of the CG formalism, based on Chein \& Mugnier specification~\cite{chein_conceptual_2008}, with a focus on elements used in most of the studied fuzzy extensions. Section~\ref{sec:axes} presents the different axes of discussion through the prism of which the fuzzy formalisms are studied. The following sections identify and discuss different inclusions of imprecise knowledge in CG: Section~\ref{sec:noeud} details the propositions that enrich concept nodes with a weighting coefficient, Section~\ref{sec:val} deals with the use of fuzzy values \add{within} concept nodes, Section~\ref{sec:dis} presents the particular case of nodes including several concepts, Section~\ref{sec:autres} discusses contributions on \add{simpler} fuzzy extensions or transpositions of previous cases.

\section{\add{Review of} conceptual graphs}\label{sec:edla}

Conceptual graphs (CG) are constituted of ontological knowledge on which factual knowledge is based. These two kinds of knowledge are successively described in this section. The ontological part \add{expresses} what can be represented, through the definition of the \emph{vocabulary}, which constitutes a terminology of knowledge. The factual part uses the elements defined in the ontological part to represent facts. All kinds of knowledge, either ontological or factual, have a logical form, not used in this paper, and a graph form, i.e. a set of labeled graphs.

Throughout the paper, we use the following illustrative example in order to provide an intuitive understanding of the defined elements as well as their usefulness: the aim is to represent the factual knowledge linguistically described by the sentence \emph{"Nouka is a student who attends course 1H001 which is a history lesson."}

\subsection{Ontological knowledge: vocabulary}
Formally, the vocabulary is defined as a quadruplet~$\mathcal{V}$ = $(T_C, T_R, M, \tau)$ whose elements are described hereinafter. 

$T_C$ and $T_R$ respectively correspond to the types of concepts and the types of relations, which respectively correspond to predicates from first-order logic of arity 1 and of arity equal \add{to} or greater than 1. They are two finite and partially ordered sets. In the case of the considered example, \emph{Student} and \emph{History} are elements of~$T_C$ whereas \emph{attend} is an element of $T_R$. We also introduce the concept \emph{Course} such that the partial order on $T_C$ gives \emph{Course} as more general than \emph{History}; \emph{Student} and the two other types are incomparable.

Each relation type has a fixed arity,  $T_R$ can be \add{partitioned} according to these arities. For instance, \emph{attend} is a relation type with arity 2.

$M=I\cup\{*\}$ is a set of markers used to instantiate the concept nodes: $I$ is the set of individual markers which identify a specific entity from the considered universe of discourse and $*$ the generic marker \add{that} references an unspecified entity. For instance \emph{Nouka} and \emph{1H001} are specific entities from $I$. 

$\tau$ is a function from $I$ to $T_C$ defining the most specific concept type instantiated by each individual marker. For instance, $\tau(\emph{Nouka}) = \emph{Student}$; $\tau(\emph{1H001}) = \emph{History}$. 

\subsection{Factual knowledge: labeled graphs}\label{subsec:edlafact}

A conceptual graph is a bipartite labeled multigraph represented by a quadruplet~$G = (C, R, E, label)$ referring to the defined vocabulary~$\mathcal{V}$.

$C$ and $R$ respectively correspond to concept nodes and relation nodes. Unlike classic multigraphs, the relations are not represented by edges but by relation nodes. $E$ is the set of edges connecting elements of~$C$ to elements of~$R$, i.e. concept nodes to relation nodes. $label$ is a labeling function from~$C$ to~$T_C\times~I$, from~$R$ to~$T_R$ and from~$E$ to the set of natural numbers~$\mathds{N}$.

\begin{figure}[t]
\centerline{\includegraphics[width=0.7\linewidth]{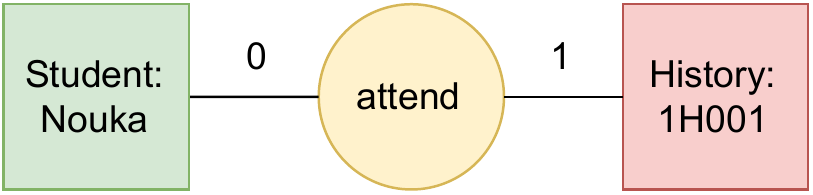}}
\caption{Representation of the considered example using the graph notation.}
\label{fig:example}
\end{figure}

The considered example can be represented by a CG \add{composed of} two concept nodes, respectively labeled~(\emph{Student}:~\emph{Nouka}) and~(\emph{History}:~\emph{1H001}), and a relation node connected to each concept node, labeled~\emph{attend}. Figure~\ref{fig:example} represents this example using a graph notation, with three nodes~$n_1$,~$n_2$ and~$n_3$, \add{where} $label(n_1) = \emph{Student}:~\emph{Nouka}$, $label(n_2) = \emph{History}:~\emph{1H001}$ and $label(n_3) = \emph{attend}$, while the edges are labeled $0$ and $1$ to specify the respective role of each concept node in the relation \emph{attend}.

The classical notation is used along the paper: a concept node labeled with type~$c$ and marker~$m$ is denoted
\begin{equation}\label{eq:node}
    [c:m]
\end{equation}
and a relation node labeled with the type~$r$, connected to two concept nodes respectively labeled with~$[c:m]$ and~$[c':m']$
\begin{eqnarray}\label{eq:gc}
& (r) & -0-[c:m]\\
 & &  -1-[c':m'] \nonumber
\end{eqnarray}
where~$0$ and~$1$ are the label of the edges identifying the role of concepts~$[c:m]$ and~$[c':m']$ within the relation~$r$.

Following these textual notations, the fact corresponding to the considered example, graphically represented in Figure~\ref{fig:example}, is denoted:
\begin{eqnarray}\label{eq:gcEx}
    &  (attend) & -0-[Student:~Nouka] \nonumber  \\
    & & -1-[History:~1H001] \nonumber
\end{eqnarray}

\section{Overview}\label{sec:axes}

This section presents the reading grid we propose to detail by which prism the papers are studied. We then give the taxonomy used to categorize the models and their respective interpretations.

We name fuzzy conceptual graph, denoted fCG, a CG enabling the representation of imprecise knowledge. Each proposition integrating fuzzy components is first presented by its definition. It is a transposition of the original paper adapted \add{to} the notation introduced in Section~\ref{subsec:edlafact} (Eq.~\ref{eq:node} and~\ref{eq:gc}). A formalization using the textual notation follows in order to summarize the proposition. 

Our proposals then consist in a critical look at the knowledge representation capacities offered by the studied models, insisting on their respective properties as well as their comparison; the study of different interpretations of these models in order to highlight their potential and their limits; and finally further details on the part of the CG impacted by the model, as well as an explicitation of the relaxed constraints.

Table~\ref{tab:taxPos} presents a taxonomy that categorizes the \add{studied} models according to which \add{type of knowledge in CG} is impacted. The section where the proposal is introduced is given in the second column. The third column contains a code, that refers to the different interpretations of the models. The first number is identical to the corresponding equation, and the rest of the code distinguishes the interpretations.

\begin{table}[t]
    \centering
    \begin{tabular}{|c|c|c|}
        \hline
        Location & Section & Code\\
        \hline
        \textbf{Ontological knowledge} & \ref{subsec:noeudVal} & (3b) \\
        &  \ref{subsec:hier} & (8b)\\
        &  \ref{subsec:regl} & (13)\\
        \hline
        \textbf{Concept type} & \ref{subsec:noeudOnt}  & (5) \\
        & \ref{subsec:disj} & (8a)\\
        & \ref{subsec:conj} & (7)\\
        \hline
        \textbf{Relation type} & \ref{subsec:rel} & (9); (10); (11)\\
        \hline
        \textbf{Individual marker} & \ref{subsec:valAtt} & (6b)\\
        \hline
        \textbf{Value} & \ref{subsec:valAtt} & (6a)\\
        \hline\hline
        \textbf{Factual knowledge} & \ref{subsec:prop}& (12)\\
        \hline
        \textbf{Concept} & \ref{subsec:noeudVal}; \ref{subsec:noeudVar} & (3a); (4)\\
        \hline
    \end{tabular}
    \caption{Taxonomy of imprecise knowledge location}
    \label{tab:taxPos}
\end{table}

As Morton~\cite{morton1987conceptual}'s works could not be accessed, they are studied through their transcript in Wuwongse and Manzano's paper~\cite{wuwongse1993fuzzy}.

\section{Weighted concept node}\label{sec:noeud}

A first family of methods proposes to represent imprecise knowledge by enriching the concept nodes. In particular, the discussed models quantify imprecision by a weight associated with the node. The first model uses a numerical value between $[0,1]$ for this purpose; the second one uses a linguistic variable, such as \emph{veryTrue}, which refers to a fuzzy set on $[0,1]$; finally the third one defers the definition of the weight associated with the node to the level of a fuzzy lattice derived from $T_C$.

\subsection{Weighting by a numerical value}\label{subsec:noeudVal}
\subsubsection{General principle}

Morton~\cite{morton1987conceptual} enables the representation of imprecise knowledge in a concept node with individual marker by associating it with a value between~$0$ and~$1$. The node takes the form:
\begin{equation}\label{eq:noeudVal}
    [c:i, \alpha], \alpha\in~[0,1]
\end{equation}
where~$c$ is a concept type in~$T_C$ and~$i$ is an individual marker in~$I$. Such a weight is not allowed for a concept node labeled with the generic marker $*$ in this model. 

The motivation is the context where a perceptual system has limitations in its ability to observe the external world. A compatibility between the considered concept type and the considered entity (represented by the individual marker) is then defined within the concept node. 

For the illustrative example, we can have for example:
\begin{equation}\label{eq:noeudValEx}
[\emph{History: 1H202}, 0.7] \nonumber
\end{equation}
The value may represent the inherent vagueness of a type, e.g. \emph{"History covers many elements that are not clearly defined, so it is difficult to say that 1H202 is a history course"}, or the difficulty of determining precisely its compatibility with the entity under consideration, e.g. \emph{"1H202 is a history course, but also contains geography parts"}.

Morton proposes definitions of the compatibility function for both the graph and the logical forms. The first one consists in the definition of a partial function~$\mu_k$ from~$T_C\times~I$ to~$[0,1]$ for each node concept~$k\in~C$. The second one consists in the definition of~$\mu:T_C\times~I\mapsto~[0,1]$ in the vocabulary, a function \add{common to} any fCG defined on that vocabulary.
The imperfect knowledge inclusion is performed at a different level depending on the considered definition. Thus according to Wuwongse and Manzano~\cite{wuwongse1993fuzzy}, as detailed below, Morton defines either the~$\alpha$ value as being specific to each individual concept~$(3a)$, or as being defined at the level of the vocabulary~$(3b)$, and thus globally.

\subsubsection{Individual weighting}

In definition~$(3a)$, the imprecise knowledge inclusion is performed \add{in} the factual knowledge, \add{in the} concept nodes, \add{and} more precisely for each node individually. This inclusion defines a fuzzy set on~$T_C\times~I$ specific to each individual concept node of a given fCG. Thus, from one individual concept to another, the same couple can have different compatibility values. One can then have~\emph{[History:~1H202,~0.7]} and~\emph{[History:~1H202,~0.4]} within the same fCG, meaning that the perceptual system that provides the information changes its conclusions about the features of~\emph{1H202} because of its own imprecision, or because of new information for example. The constraint thus relaxed by this fuzzy integration model is the degree to which an entity belongs to a concept type.

We propose two readings of the constraint relaxation. According to the first one, the fCG keeps the classical constraint: only one couple of~$T_C\times~I$ is associated to each individual concept node~$(3a1)$. In this case the fuzzy set of a given node is defined only for a single pair of~$T_C\times~I$, which makes the definition of~$\mu_k$ excessive.

In the second reading, we consider that this constraint is also relaxed, which leads to fCG whose individual concept nodes are fuzzy sets on~$T_C\times~I$, modeling  imprecision both on the concept type and on the individual marker~$(3a2)$. Thus, one can have the node~\emph{[(History: 1H202, 0.7),(Geography: 1H202, 0.4)]}, which can represent imprecision on several type-entity pairs. Section~\ref{sec:dis} more generally discusses such nodes labeled by several pairs of $T_C\times~I$, and examines several interpretations, in particular conjunctive and disjunctive.

These fuzzy sets remain specific to each individual concept. For~$(3a2)$ as for~$(3a1)$, following the definition provided by Morton~\cite{morton1987conceptual} according to Wuwongse and Manzano~\cite{wuwongse1993fuzzy}, this can be interpreted as the limitations of a perceptual system that \add{is not} static. For each individual concept, the perceptual system has a \add{level of} precision on its observation depending on the context (or other parameters). This precision can then vary for the same given entity from capture to capture.

\subsubsection{Global weighting}
In definition~$(3b)$, the imprecise knowledge is included at the ontological level, in the vocabulary. Thus we propose the interpretation that compatibility becomes inherent of every pair of~$T_C\times~I$, regardless of the node~$k\in~C$ under consideration, because it is not established during the observation of each entity in~$I$. The compatibility that was defined for the triplet~$(k, T_C, I)$, for~$k\in~C$ of a given CG, becomes here defined for the triplet~$(\mathcal{V}, T_C, I)$. 
 This interpretation may thus refer straightforwardly to the inherently vague character of a type. 

There is therefore the guarantee that, for instance, any occurrence of the node~\emph{[History: 1H202, 0.7]} in any fCG of the same vocabulary, its weight is always \add{the} same. 

The differences with case~$(3a)$ have various possible motivations: the origin of the imperfect piece of information, the paradigm of capturing this piece of information or its interpretation in imperfect knowledge. The first motivation expresses that cases~$(3a)$ and~$(3b)$ are different because the imperfect piece of information stems from different causes: in case~$(3a)$ the imperfect knowledge defined in the factual part of the fCG is justified by its production by an imperfect perceptual system. In case~$(3b)$ the anchoring in the ontological part is justified by the partial compatibility inherent to each type-entity couple.

The second motivation translates the fact that in case~$(3b)$ there is a unique compatibility value for a given type-entity couple, whereas with case~$(3a)$ there is a multitude of values for the same couple. This multiplicity of values for the same couple can be justified by captures spread out in time, whose parameters of observations would vary, or by the capture by different imperfect systems, with diverse characteristics.

Finally, a third motivation is a difference in the interpretation of the collected pieces of imperfect information, either in terms of uncertain or vague knowledge, as proposed by the author, but also in terms of imprecise knowledge or level of confidence in the knowledge. Case~$(3a)$ describes a weight that may vary between the observations of a given type-entity couple. It is in line with the uncertainty or the level of confidence interpretations, as there are variations of levels of confidence or certainty associated with the observations of a given type-entity couple. Case~$(3b)$ describes a weight inherent to the type-entity couple. It is in line with the imprecise or vague knowledge interpretation, as there may be a level of imprecision or vagueness associated with a given type-entity couple, which is specific to this couple and does not vary over the different observations of the same couple.

\subsection{Weighting by a linguistic variable}\label{subsec:noeudVar}

A second model, proposed by Wuwongse and Tru~\cite{wuwongse1996towards}, considers weights in the form of a linguistic variable~$\lambda$ associated with a fuzzy set on~$[0,1]$. {A concept node is written}:
\begin{equation}\label{eq:nodeVar}
     [c:m, \lambda], \lambda\in K
\end{equation}
where~$m\in M$ and~$K$ is a set of linguistic variables defining the degree of truthfulness, e.g.~$K$~=~\{\emph{veryTrue},~\emph{true},~$\ldots$~\}. Each of the terms is associated with a fuzzy set defined on the universe~$[0,1]$, instead of a single value~$\alpha\in~[0,1]$ for Morton.

Considering the illustrative example, one can have \emph{[History:~2H132,~quiteTrue]} and~\emph{[Geography:~2H132,~veryFalse]}. 
This example represents \emph{"It is quite true that 2H132 deals with history and it is very false that 2H132 deals with geography"}. The values~\emph{quiteTrue} and~\emph{veryFalse} are defined at the ontological level and are therefore not specific to an individual node.

This model uses a linguistic variable instead of a numerical value which can be considered as making the manipulation of imprecision by humans more intuitive, following a usual principle in fuzzy logic.

According to Wuwongse and Tru~\cite{wuwongse1996towards}, this choice leads to more relevant results when integrated in the projection operation. However their arguments seem debatable and the claimed issue can be related to differences of interpretation with respect to Morton's framework.

The fuzzy sets associated with the linguistic variables~$\lambda$ are elements of the set~$K$ defined at the level of the vocabulary, which potentially allows the definition of an infinity of variables. This moderates the previous statement about the more intuitive use of linguistic variables.

The set~$K$ is partitioned into three sets~$T$, $F$ and~$U$ referring to~"True", "False" and~"Unknown" respectively. The membership of a linguistic variable in~$T$, $F$ or~$U$ gives properties on this variable: terms in~$T$ are associated with increasing membership functions on~$[0,1]$ and equal to~$1$ on~$1$; terms in~$F$ are associated with decreasing membership functions on~$[0,1]$ and equal to~$1$ on~$0$.

This partition alters the reasoning mechanisms. Indeed, it implies that inference between types of concepts associated with terms of~$K$ is not directly derived from the classical inference principles between basic concept types and between fuzzy sets: it is defined to take into account the type of concept, the degree of truth and the type of truth value.

While Morton's proposal concerns individual markers only, Wuwongse and Tru's allows the use of the generic marker~$*$ as well; it enables the representation of a compatibility between a type and an undefined entity. This highlights the fact that the imprecision represented is not inherent of pairs in~$T_C\times~I$, because then the linguistic value associated with a generic node, \emph{[History: *, true]} for example, becomes uniquely defined as a function of type.

This second definition of fuzzy node relaxes the constraint at the level of the factual knowledge of CG, in a similar way as~$(3a)$, i.e. at the level of the compatibility between the entity of a concept node and its type, but also allows the use of generic marker and the use of fuzzy sets instead of a single numerical value. 

Thus, Wuwongse and Tru also relax the constraint on the precision of the represented entity (it can be less precise through the use of the generic marker) and on the precision of the compatibility values~$(4)$.

\subsection{Ontological weighting}\label{subsec:noeudOnt}

A third model, proposed by Cao~\cite{cao1997fuzzy}, does not represent fuzzy types at the level of factual knowledge, but considers types belonging to a lattice of fuzzy types, giving for~$m\in~M$ and~$c_\lambda$ in this lattice:
\begin{equation}\label{eq:nodeTyp}
    [c_\lambda:m], \lambda\in~K
\end{equation}

In the case of the considered example, one can for instance build the concept node~$\emph{[History}_\emph{veryTrue}\emph{: 2H132}]$.

A fCG using such fuzzy concept types 
is highly similar to a classical CG. Indeed, the concept nodes are type-marker pairs whose type belongs to a set with a partial order relation. The variable $\lambda$ in Eq.~(\ref{eq:nodeTyp}) is not determined during the construction of the factual node, but in the vocabulary, and it is associated with a type. One then has access to different occurrences of the basic type \emph{History} associated with different values of $\lambda$, resulting in different fuzzy types. 

By this proposition, as with~$(3b)$, the weighting is explicitly linked to the types, and the imprecision is located at the level of the ontological knowledge~$(5)$.

\section{Imprecise attribute value}\label{sec:val}

This section discusses a different case of fuzzy semantic integration: fuzziness occurs within a concept node as well, but for a particular type of concept, called concept with attribute. Such concept nodes contain an attribute value in addition to, or instead of, a marker. The fuzzy extension of such nodes allows the definition of linguistic variables for the attribute value. After recalling the definition of these concept types in the crisp case, this section successively discusses several fuzzy extensions proposed in the literature. The second one is formulated as
fuzzification of markers, however it represents fuzzy values.

\subsection{Concept type with attribute}\label{subsec:valAtt}
In the classical CG formalism, concept types with attribute~\cite{chein_conceptual_2008} constitute a type of concept associated with a value domain~$U(c)$. A concept node with attribute is denoted~$(c:v)$ with~$c$ the type of concept attribute and~$v$ a value in~$U(c)$.

For example, one can represent~\emph{[Grade: 90]}, where~\emph{Grade} is a concept type with attribute domain~$U(Grade) = [0,100]$ and~$90$ is a value in the domain~\emph{U(Grade)}.

In Sowa's formalism~\cite{sowa_conceptual_1983}, the values are symbols, and the markers of the concept nodes with attribute remain, and by default has a value of 1. Buche et al.~\cite{buche2001representation} on the other hand introduce the type~\emph{Value} which is linked to a relation node of type~\emph{Val} linked to a node with 
of this type are linked to the parameter considered by the node relationship~\emph{Val}. Such concept nodes with attributes can actually be represented in the classic formalisms of CG, as recalled Section~\ref{sec:edla}: they can be represented as nodes linked through a~\emph{Val} relation node to specific concept nodes with \emph{Value} as type and the attribute value as marker.

Types with attributes can thus be formalised in different forms. In one case, a concept node with attribute is modeled as a triplet type-marker-value, while in the other case it is modeled as a couple type-marker linked through a relation node of type \emph{Val} to a \emph{Value}-value couple. The concept node with attribute in the example are not based on either model and are a simplification to illustrate our point, with such a node ending as a couple type-value. It is the basis for concept node with attribute notation in following sections.

\subsection{Linguistic variable as a value}\label{subsec:valVar}

\subsubsection{General principle}

Morton~\cite{morton1987conceptual}, according to Wuwongse and Manzano~\cite{wuwongse1993fuzzy}, enables the representation of imperfect knowledge in the value field of a concept node having a concept type with metric attribute. The latter corresponds to the case where the attribute value has an associated measure.

The imperfect knowledge is represented by a linguistic variable~$\lambda$ as a fuzzy set over~$U(c)$, the value domain of the concept type with metric attribute $c$:
\begin{equation}\label{eq:val}
    [c:\lambda], \lambda\in~\mathcal{F}(U(c))
\end{equation}
where~$\mathcal{F}(U)$ denotes the set of fuzzy sets defined on domain~$U$.

For example, if~\emph{Grade} is metric, one can construct the node~\emph{[Grade: Good]} where~\emph{Good} is a linguistic variable associated with a fuzzy set on~$[0,100]$, whose value is determined by the measure associated with the concept type with metric attribute~\emph{Grade}.

Wuwongse and Manzano~\cite{wuwongse1993fuzzy} extend this definition to the non-metric case, whose  distinction with the metric case is open to interpretation. The authors specify for example, for a metric attribute concept type~\emph{Size} and a continuous universe~$U$, that~$U(Size)$ is a continuous set corresponding to the measure associated to~\emph{Size}. For the non-metric case, it is a discrete set~$U'$. They illustrate the latter with a concept~\emph{Color} and  an associated categorical value~\emph{Red}, which is not a value defined by a measure. In both cases the associated values are fuzzy sets on~$U(Size)$ and~$U'$ respectively. The distinction seems to inform on the formalization or not of a measure function associated to the type of concept with attribute. In the case where no measure is associated, the set~$U'$ thus corresponds to a discrete universe of categories associated to the concept type with non-metric attribute.~$(6a)$

\subsubsection{Fuzzy marker}

A definition similar to the previous one introduces the notion of~\emph{Fuzzy Marker}, based on Buche et al.'s value model~\cite{buche2001representation}, by Thomopoulos et al.~\cite{thomopoulos2003representation}.  It differs in that the domain of any concept type with attribute is defined on~$I$, the set of individual markers, instead of a universe~$U$ distinct from~$I$. In these models, a fuzzy marker is a fuzzy set on~$I$ restricted to the domain of the associated concept type~$t_c$ within the concept node~$c$. A classical marker is the special case of a crisp set that associates~$1$ as the degree of membership to the considered element~$m$ of~$I$ on the domain of~$t_c$, and~$0$ to the others.

It allows the representation of imprecision by going from a single precise value to a set of weighted values. For each metric attribute type~$c$ a membership function~$\mu_c$ is defined on~$U(c)$, i.e. subset of the domain of all possible values corresponding to the value domain of $c$. In the non-metric case, it is defined on a discrete domain~$U'$.~$(6b)$

\subsubsection{Analysis}

For both definitions, the notion of fuzzy attribute type enables the use of a linguistic variable on the value domain.

In Thomopoulos et al.'s definition, there is a confusion between the individual marker~$i$, which refers to an entity in reality, and a value~$v$, which corresponds to the measure of a characteristic. As a result, syntactically this model does not distinguish between concept types with attributes and classical concept types; however, the semantics remain different, as these two kinds of concept types have a different interpretation in reality: a value of an attribute and a symbolic marker referring an entity in reality are two different things, so the confusion within this model persists. It should be noted though that in the application context presented by Thomopoulos et al.~\cite{thomopoulos2003representation}, where CGs are used to represent data in microbiology, the confusion may never happen in practice.

On another note, the distinction between metric and non-metric attribute types may have its origin in the fact that~$U$ is usually the universe on which concepts, relations and individual markers are interpreted in logical semantics~\cite{chein_conceptual_2008}. 

In the definition of fuzzy markers, the use of~$I$ allows the definition of domains of values on~$I$, discrete or not, continuous or discontinuous, which is a second modification of this set, in addition to the fact that its elements no longer necessarily refer to entities, but both entities and attribute values. Or, rather, attribute values are then considered as entities.

It is not clear whether Morton, Wuwongse and Tru's definitions define an imperfect knowledge inclusion on the factual or ontological side. On the one hand there is the possibility that the value domain~$U(c)$ of a metric attribute type~$t$ points to a fuzzy set, thus globally defined, which is in line with an inclusion on the ontological part. However, this possibility is not made explicit before its use in a CG, because no additional element is defined in the ontological part. It is indeed not explicit that linguistic variables defined on~$U(c)$ are defined prior to their use in the factual part. In the case of the fuzzy marker, the definition is clearly on the ontological part with respect to the value domains, and on the factual part with respect to the fuzzy sets on these value domains, since they are not associated with any variable or element of the support.

\section{Fuzzy multi-concept node}\label{sec:dis}

A third type of imperfect knowledge integration is observed in the case of multi-concept nodes, which classically correspond to conjunctive nodes as recalled in Section~\ref{subsec:rap} below. Three fuzzy variants are then presented successively.

\subsection{Conjunctive type introduction}\label{subsec:rap}

Multi-concept nodes extend classical CG enabling the use of conjunctive types~\cite{cao1999foundations,baget2003simple,chein_conceptual_2008}: the node labeling function associates concept nodes with a subset of~$T_C$ of incomparable types. Two types are said to be incomparable if they cannot be compared by the partial order relation defined on~$T_C$, i.e. one is not the generalization of the other and vice versa. Thus the \emph{label} function of a CG does not associate each node~$k\in~C$  with a single element of~$T_C$, but with a subset of~$T_C$ . 

It has to be noted that the conjunctive type does not correspond to an increase in expressiveness but to a syntactic shortcut in the factual part of CG: a node with a conjunctive type~$c$ consisting of~$n$ incomparable types can be replaced by $n$ nodes whose respective types are the $n$ types constituting the conjunctive type~$c$. The conditions to obtain equivalence are that each resulting node has the same connectivity as the multi-concept node, and that each node refers to the same entity.

\subsection{Fuzzy conjunction of types}\label{subsec:conj}

Cao~\cite{cao1999foundations} proposes a natural extension to the fuzzy case, by replacing the considered concepts by fuzzy concepts as presented in Section~\ref{subsec:noeudOnt}.  It is a conjunction of such fuzzy concepts, and one obtains with~$c$ and~$c'$ in~$T_C$, $\lambda$ and~$\lambda'$ in~$K$, $m$ in~$M$: 
\begin{equation}
     [c_\lambda, c'_{\lambda'}: m]
\end{equation}

In the same way as for the conjunctive type, it is a syntactic shortcut which does not increase expressivity as compared to the propositions on nodes and fuzzy types.

For example, one can build \emph{[History$_{true}$, Geography$_{notTrue}$~:~1H002]} where \emph{1H002} can be a history-geography lesson presenting only a few maps to illustrate the \add{geography} part. Note that in this example, the conjunction of fuzzy types used is relevant, it makes sense conceptually, but it is not always the case. For instance, the conjunction of \emph{Student} and \emph{Course} is not conceptually clear, even though it can occur and be relevant in practice. For instance, it can be the case of a student both attending some courses and being the subject of one \add{course} in particular.

\subsection{Fuzzy disjunction of types}\label{subsec:disj}

In a different fuzzy multi-concept case, Thomopoulos et al.~\cite{thomopoulos2003representation} propose to define fuzzy disjunction of incomparable types, giving for~$c$ and~$c'$ two types, and~$\alpha$ and~$\beta$ two values on~$[0,1]$:
\begin{equation}\label{eq:disj}
    [(c, \alpha), (c', \beta): *]
\end{equation}

The context of this proposition is the use of a CG as queries in databases, where the results have to match the query CG, i.e. be equal to it or be a specialization of it. Then the defined disjunction corresponds to an imprecision or a preference on the nature of the sought information, and as such enables more expressive queries returning a broader range of results. The results must correspond to the disjunctive type, i.e. be of the same type or a subtype, in proportion to the importance of the weight associated with each type. Whether the imprecision or preference interpretation is chosen modifies the notion of distance with the query CG, and thus returns different results. In this framework, the weight is a generic marker associated to a fuzzy set on~$I$ corresponding to all individual markers matching the fuzzy disjunction of types, where each of them has a membership degree equal to the matching type's degree in the fuzzy disjunction.

With this definition, one can for example build~\emph{[(History, 0.8),~(Geography, 0.4): *)]} which can represent the lessons that~\emph{Nouka} would prefer to have next year.

Compared to the previous definitions of crisp and fuzzy conjunction of types, it must be underlined that the disjunction 
does not reduce to a syntactic shortcut. Indeed, the equivalence between a multi-concept node in the conjunctive case and several nodes is not possible in the disjunctive case, because there is, to our knowledge, no other knowledge disjunction representation in the CG formalism.

The relaxed constraint is the degree of truth associated with each type. Moreover, an additional notion of fuzziness is added with the relaxation of the constraint on the type uniqueness. In mono-concept types cases, a node is associated with a unique type, even in the fuzzy type case, so that one can state that "this entity of such type", with a possible degree of truth associated. Here one cannot state that the represented entity is of such or such type, hence the additional relaxation of constraint. We thus have a fuzzy set on a subset of incomparable types of~$T_C$ instead of a single element of~$T_C$.~$(8a)$

\subsection{Fuzzy hierarchy}\label{subsec:hier}

Thomopoulos et al.~\cite{thomopoulos2003different} extend the previous definition by deriving from the fuzzy disjunction of type present in the query, weights for all types of~$T_C$, i.e. even the ones not present in the disjunction. It is a so-called~\emph{developed} form named \emph{fuzzy type in extension} in the paper proposing it, in opposition to the fuzzy disjunction of types named then \emph{fuzzy type in intention}. Depending on whether the interpretation is preference or vagueness,  respectively the maximum or minimum of the supertypes is recursively chosen for each subtype, and 0 for the more general types, as illustrated in the example below.

This is a fuzzy type definition that is specific to the query case in a CG database presented by the paper. In this sense, other proposals~\cite{thomopoulos2006fuzzy} impose to systematically take the maximum in order not to lose any possibly relevant result, whichever the chosen interpretation.

For instance, considering the previous the example of the previous definition~\emph{[(History,~0.8),(Geography,~0.4):~*)]}, one can build the hierarchy~$T_C =$[(\emph{History},~0.8), (\emph{Geography},~0.4), (\emph{Course},~0), (\emph{History-Geography}, $\alpha$)], 
where~\emph{Course} is more general than~\emph{History} and~\emph{Geography} and \emph{History-Geography} is more specific than~\emph{History} and~\emph{Geography}. The $\alpha$ coefficient takes value~$0.8$ in the case of a preference interpretation, and~$0.4$ in the case of a precision interpretation. 

This definition has the particularity of including its own interpretation: the definition depends on the interpretation, contrary to the previously discussed ones. Indeed, the latter offer definitions in possibilities of interpretations, and treat the influence of this interpretation only in the reasoning inference step. It makes explicit the interpretation of what a fuzzy disjunction of type on~$T_C$ represents.~$(8b)$

\section{Other fuzzifications}\label{sec:autres}

This section deals with other cases of imperfect knowledge inclusion which do not lead to much discussion, either because of their simplicity or because they are only a transposition of previously detailed cases. The case of fuzzy relations is first discussed, then the case of propositional fuzziness and finally the case of fuzzy inference rules.

\subsection{Fuzzy relations}\label{subsec:rel}

The integration of fuzzy components can apply to the relation types, beyond the concept types discussed in Section~\ref{sec:noeud}, in a straightforward way and following the same lines of discussion. Indeed, the weights associated with the concepts described in Section~\ref{sec:noeud}, whether they are numerical or linguistic, can be naturally extended to relations.

\subsubsection{Weight as numerical value}

Wuwongse and Manzano~\cite{wuwongse1993fuzzy} propose to generalize the principle discussed in Section~\ref{subsec:noeudVal} for the case of concept nodes to the case of relation nodes: Eq.~(\ref{eq:gc}) can be enriched with a numerical weight~$\alpha\in [0,1]$, leading to:
\begin{eqnarray}\label{eq:relVal}
& (r, \alpha) & -0-[c:i]\\
 & &  -1-[c':i'] \nonumber
\end{eqnarray}
The weight can be interpreted as representing the compatibility of the nodes concepts~$c$ and~$c'$ defined by a function specific to the relation~$r$ returning~$\alpha$. 

For the considered example, one can for instance build:
\begin{eqnarray}
& (\emph{attend}, 0.8) & -0-[\emph{Student: Nouka}] \nonumber\\
&  & -1-[\emph{History: 1H008}] \nonumber
\end{eqnarray}
which represents the fact that the student~\emph{Nouka} attended most of the history class~\emph{1H008}.

\subsubsection{Weight as linguistic value}
Similarly, Wuwongse and Tru~\cite{wuwongse1996towards} propose to apply the principle discussed in Section~\ref{subsec:noeudVar} for concept nodes to the case of relation nodes, i.e. to weight relation nodes with a linguistic variable defined on~$K$:
\begin{eqnarray}\label{eq:relVar}
& (r, \lambda) & -0-[c:i]\\
 & & -1-[c':i'] \nonumber
\end{eqnarray}
where each linguistic variable $\lambda\in~K$ corresponds to a fuzzy set on~$[0,1]$.  


\subsubsection{Weight at the ontological level}

Finally, the case of weights at the ontological level formalised by a fuzzy lattice, as presented in Section~\ref{subsec:noeudOnt} for concept types, is applied to the case of relation types by Cao et al.~\cite{cao1997fuzzy}:
\begin{eqnarray}\label{eq:relOnt}
& (r_\lambda) & -0-[c:i]\\
 & & -1-[c':i'] \nonumber
\end{eqnarray}
where $\lambda$ is a linguistic variable in~$K$ as introduced earlier. 

From the point of view of the fCG, the relation~$r_\lambda$ is a classical type of relation belonging to a particular lattice of types, a lattice of basic types associated to linguistic variables in~$K$. In a sense, as in the case of concepts, such a fCG is a classical CG with a particular vocabulary.

\subsection{Weighting a concept node description}\label{subsec:prop}

Morton~\cite{morton1987conceptual}, according to Wuwongse and Manzano~\cite{wuwongse1993fuzzy}, uses a notion of concept node triplets consisting of a type~$c$, a marker~$m$ and a description~$d$. This description is itself a CG that describes the concept node. This notion is available in Sowa's formalism~\cite{sowa_conceptual_1983}, and is similar to that of nested CGs~\cite{chein_conceptual_2008}, restricted to a single level of nesting. 

Morton generalises this formalization of CGs and includes a weight of the description~$d$ which is written:
\begin{equation}
    [c:m, (d, \alpha)]
\end{equation}

The value~$\alpha$ represents the compatibility between~$c$ and~$d$, and is, as the other Morton definitions, modeled by a compatibility function. It is specified that compatibility can also be modeled by a function from~$[0,1]$ to~$[0,1]$ representing then a compatibility with truth values in~$[0,1]$. This compatibility only affects the factual part of CGs as it is defined for each couple type-description, and that the descriptions in themselves are CG which are not, apriori, defined in the ontological part.

\subsection{Fuzzy rule}\label{subsec:regl}

A $\lambda$-rule in the CG formalism, also called inference rule,  is a classical extension of the CG reminded in Section~\ref{sec:edla}, allowing to perform reasoning inference~\cite{chein_conceptual_2008}:
for two concepts~$c$ and~$c'$, a relation~$r$, an individual marker~$i$ and a variable $*x$, such a rule can be written:
\begin{equation}\label{eq:rule}
    [c:*x] \Rightarrow [c:*x]-1-(r)-0-[c':i]
\end{equation}
It is interpreted as a rule of the form \emph{IF a subset of the CG matches the hypothesis (e.g. in Eq.~(\ref{eq:rule}) the considered CG contains~\emph{$x$ of type~$c$}), THEN the CG can be extended and/or specialized to match the conclusion (e.g. in Eq.~(\ref{eq:rule}) \emph{$x$ is in relation to~$i$ of type~$c'$ via the relation~$r$})}. Rules with more complex premises and conclusions can be considered. 

For example, we may have the rule:\\
\emph{[History: *x]} $\Rightarrow$
    \emph{[History: *x]-1-(learned)-0-[Student: *]}
    
which represents the piece of knowledge \emph{If~$x$ is a history lesson, then there exists a student who learns  lesson~$x$}.

Wuwongse and Tru~\cite{wuwongse1996towards} introduce fCG programs and Cao and Creasy~\cite{CAO2000Fuzzy} the \emph{expansion} of $\lambda$-fCG and universally quantified fCG, which uses universal quantifiers instead generic markers. These propositions can be used as extensions of the $\lambda$-rules so that they allow for fuzzy weights in the premise and in the conclusion.

In the first case~\cite{wuwongse1996towards}, they are the direct transposition of the $\lambda$-rules to the case of fCG having fuzzy types at the level of concepts and relations, and fuzzy values at the level of the value of an attribute type. Thus the hypothesis and the conclusion of the rule are such graphs, and the specific reasoning mechanisms that this induces are detailed. These specific inferences are a transposition of the deduction rules defined for  fCG
~\cite{wuwongse1996towards}.

In the second case~\cite{CAO2000Fuzzy}, the operation of \emph{expanding} the $\lambda$-fCG and the universally quantified fCG puts these specific fCG in a different form, that corresponds to a $\lambda$-rule. For example,  considering the universally quantified CG:
\begin{eqnarray}
& (\emph{attend}) & -0-[\emph{Student:} \forall] \nonumber\\
 & & -1-[\emph{History: 1H003}] \nonumber
\end{eqnarray}
which represents \emph{All the students attended the history class 1H003}, its extension, which corresponds as stated above to a $\lambda$-rule, is written:
\begin{eqnarray}
[\emph{Student:} *x] & \Rightarrow \nonumber\\
& (\emph{attend}) & -0-[\emph{Student: *x} ] \nonumber\\
 & & -1-[\emph{History: 1H003}] \nonumber
\end{eqnarray}
which represents the same information under another form. The fuzzy case is then when the quantifier~$\forall$ is for example replaced by a fuzzy set on $[0,1]$, representing a generic quantifier. 
This value represents how far one can deduce the conclusion from the represented premise, for example with the value~\emph{most} which would give \emph{Most students attended the history class 1H003}.

\section{Conclusion}

The comparative discussion presented in this paper shows the richness and diversity of the propositions allowing to model imprecise knowledge in the framework of fuzzy conceptual graphs: the latter enriches the classical conceptual graph model and increases their expressiveness integrating fuzzy components at different levels with various interpretations and uses. As surveyed in this paper, they include fuzzy concept nodes, fuzzy relation nodes, fuzzy types, fuzzy markers and fuzzy values,  as well as fuzzy inference rules. A complementary view is offered in Table~\ref{tab:taxPos} that proposes a taxonomy of the imprecise knowledge locations, focusing on the distinction between the ontological and factual parts, detailing the particularly rich variations of the ontological fuzzy extensions. It highlights the distinctions operated in this paper to formalize all proposed fCG models and discuss the various interpretations of these models, bringing to light that beyond the specific cases for which they have been defined, they offer original tools to represent imprecise information. \add{It has to be noted that these models can be combined, but a balance has to be established between the expressivity and the complexity of the resulting formalism.}

Ongoing works aim at studying the consequences of these choices on the reasoning inference processes proposed by the papers that introduced these fuzzy conceptual graph extensions, depending on the chosen interpretation that can vary, as discussed in the paper. The case of uncertain knowledge, that can rely on the same formalization as imprecise knowledge discussed in this paper but is associated with other interpretation and reasoning tools, is of specific interest.

\bibliographystyle{IEEEtran}
\bibliography{bib}

\end{document}